\documentclass[twoside,11pt]{article}

\usepackage{jmlr2e}

\usepackage{amsthm,amssymb}
\usepackage{amsmath}
\usepackage{algorithm2e}
\usepackage{graphicx}
\usepackage{todonotes}
\usepackage{subcaption}
\usepackage{multirow}
\usepackage{tikz}
\usepackage{arydshln}
\usepackage{lineno}

\usepackage{lastpage}

\ShortHeadings{Supervised Linear Centroid-Encoder}{Ghosh and Kirby}
\firstpageno{1}

\begin{document}

\title{Yet Another Algorithm for Supervised Principal Component Analysis: Supervised Linear Centroid-Encoder}

\author{\name Tomojit Ghosh \email Tomojit.Ghosh@colostate.edu \\
       \addr Department of Mathematics\\
       Colorado State University\\
       Fort Collins, CO 80523 ,USA
       \AND
       \name Michael Kirby \email Kirby@math.colostate.edu \\
       \addr Department of Mathematics\\
       Colorado State University\\
       Fort Collins, CO 80523, USA}
\editor{TBD}
\maketitle

\begin{abstract}
We propose a new supervised dimensionality reduction technique called Supervised Linear Centroid-Encoder (SLCE), a linear counterpart of the nonlinear Centroid-Encoder (CE) \citep{ghosh2022supervised}. SLCE works by mapping the samples of a class to its class centroid using a linear transformation. The transformation is a projection that reconstructs a point such that its distance from the corresponding class centroid, i.e., centroid-reconstruction loss, is minimized in the ambient space. We derive a closed-form solution using an eigendecomposition of a symmetric matrix. We did a detailed analysis and presented some crucial mathematical properties of the proposed approach.
We establish a connection between the eigenvalues and the centroid-reconstruction loss. In contrast to Principal Component Analysis (PCA) which reconstructs a sample in the ambient space, the transformation of SLCE uses the instances of a class to rebuild the corresponding class centroid. Therefore the proposed method can be considered a form of supervised PCA. Experimental results show the performance advantage of SLCE over other supervised methods.
\end{abstract}



\section{Introduction}
\label{intro}
Historically, dimensionality reduction (DR) is an
integral component of the machine learning workflow for high-dimensional data sets; see, e.g., ~\citep{dudaHart1973,kirby2001geometric,chep}. 
Recent technological advancements in data acquisition, 
and storage along with the wider accessibility of High Performance Computing (HPC), have increased the need for a broad range of tools capable of high-dimensional data  analytics~\citep{hayden2015genome}. 
For example, bioinformaticians  seek to understand {\it omics} data such as the gene expression levels measured by microarrays or next-generation sequencing techniques where samples consist of 20,000-50,000 measurements \citep{reuter2015high}. Often these high-dimensional features may be noisy, redundant, missing, or irrelevant \citep{jing2015stratified}, which has the potential to degrade the performance of machine learning tasks \citep{shen2022classification}.
Further, the number of samples 
available  is often so small as to preclude the quality training of nonlinear methods~\citep{ghosh2022supervised,aminian2021early}.

In general, a DR technique is applied as a pre-processing step to reduce data dimension to facilitate
visualization, clustering, or classification \citep{li2021novel}. 
The requirements of a DR technique typically include preserving one or more of the intrinsic properties of interest in the embedding space. The intrinsic property can be statistical \citep{jolliffe_1986,van2008visualizing},  topological \citep{Kohonen93}, or geometrical \citep{TeSiLa00,becht2019dimensionality}; in the presence of labels, the intrinsic property may consist of discriminating features to assign class membership \citep{fisher36lda}.

Traditionally, principal component analysis \citep{jolliffe_1986} (PCA) is the most widely used DR technique, a projection method that minimizes reconstruction loss. Statistically, the projection is built to capture the maximum variance from the sample covariance matrix. Despite its easy implementation and simple geometric interpretation, PCA often produces ambiguous results, sometimes collapsing distinct classes into overlapping regions in the setting where class labels are available. 
This is of course due to the 
fact PCA does not explicitly
exploit label information.
In contrast, Linear Discriminant Analysis  (LDA) uses class labels to avoid class overlap by maximizing the class separation and minimizing the class scatter, thus creating a better embedding than PCA \citep{duda2006pattern}. This motivates us (and others) to find
a supervised PCA algorithm in our quest to 
obtain the best of both worlds.

It is, in general, possible to add labels to unsupervised methods to create their supervised analogs. A heuristic-based supervised PCA model first selects important features by calculating correlation with the response variable and then applies standard PCA of the chosen feature set \citep{doi:10.1198/016214505000000628}.
Another supervised PCA technique, proposed by \citep{Barshan:2011:SPC:1950989.1951174}, uses the Hilbert-Schmidt independence criterion to compute the principal components which have maximum dependence on the labels. The Supervised Probabilistic PCA also uses the class labels to create the low-dimensional embedding~\citep{yu2006supervised} Note that, all these supervised formulations of PCA work better than the unsupervised one for data where labels are available.

In this paper, we propose a linear dimensionality reduction technique that explicitly utilizes the class labels to create a low-dimensional embedding. This algorithm is effectively a linearization of the nonlinear Centroid-Encoder \citep{ghosh2022supervised}. Being a linear model, SLCE is less prone to overfit high-dimensional data sets, requires less data for learning, and at the same time is easy to interpret geometrically. Here we summarize the main contribution of our work.
\begin{itemize}
    \item We have proposed a linear dimensionality reduction technique called Supervised Linear Centroid-Encoder (SLCE) that uses class label information. The proposed method doesn't use the response variable; instead uses the class centroid to impose supervision in learning.
    
    \item We have shown the connection between PCA and SLCE and argued that SLCE is a form of supervised PCA.
    
    \item Unlike Centroid-Encoder, which uses nonlinear mapping, the proposed technique creates a projection by mapping a sample to its class centroid.
    
    \item We proposed a closed-form solution using the eigendecomposition of a symmetric matrix.

    \item {
    We provide upper bound on the number of eigenvectors with positive eigenvalues that comprise the SLCE embedding.}
    
    \item We have shown how the eigenvalues are connected to the final objective, i.e., the centroid-reconstruction loss.
    
\end{itemize}
This paper is organized as follows: In Section \ref{lit_review}, we review the related literature on supervised DR techniques. In Section \ref{LCE}, we present Supervised Linear Centroid-Encoder including its derivation and properties. Section \ref{viz_clf_result} presents comparative experimental results on five bench-marking data sets.  We conclude in Section \ref{conc_future_work}

\section{Related Work}
\label{lit_review}
Dimensionality reduction has a long history and is still an active area of research, see. e.g., ~\citep{chepushtanova2020dimensionality,van2009dimensionality} and references therein. A variety techniques with a spectrum of optimization problems and heuristics have been discovered over the past decades.  However, considering that our proposed method is supervised and linear, we will briefly describe this class of algorithms.

Fisher's linear discriminant analysis (LDA) is historically one of the most widely used supervised dimensionality reduction techniques \citep{fisher36lda,dudaHart1973}. LDA reduces the dimension by minimizing the class scatter and maximizing the class separation in the reduced space. LDA creates the mapping for a $C$ class data using a $C-1$-dimensional sub-space. Although PCA is an unsupervised technique, several attempts have been made to incorporate the label information into the model. One such method is Bair's supervised principal component (Bair's SPCA) which is a heuristic-based approach. It's similar to PCA but uses a subset of features that have the maximum dependencies on the response variables, i.e., class label. The feature dependence on the class label is calculated by the standard regression coefficient, which is defined below 

\begin{equation}
\begin{aligned}
w_j = \frac{x_j^Ty}{\sqrt{x_j^Tx_j}}
\end{aligned}
\label{equation:reg_coeff}
\end{equation}
where $x_j$ is the $j^{th}$ variable and $y$ is the response variable. After calculating the $w_j$ or the importance of each feature, a threshold $\theta$ is used to select the most important ones, and at last,  standard PCA is applied to the selected features. The proposed method is a two-step process, and the authors used cross-validation to pick an optimal $\theta$. Notice the $\theta$ is data set dependent, and searching for an optimum value using cross-validation is computationally expensive. The shortcomings were addressed by Piironen et al., who proposed an iterative supervised principal component (ISPC) \citep{piironen2018iterative}. ISPC doesn't use cross-validation for feature screening and can be used for multiple classes.

Barshan et al. \citep{Barshan:2011:SPC:1950989.1951174} formulated a supervised PCA using reproducing Kernel Hilbert Space to maximize the dependency of a sample in the low-dimensional space on the outcome measurement. Let there be $n$ $p$-dimensional samples stacked in a $p \times n$ data matrix $X$, and $Y$ be a matrix of outcome measurement. Given a transformation matrix $U$, the model finds the solution by maximizing the dependency of the projected data $U^TX$ to Y. The kernel Hilbert Space measures the dependence between $U^TX$ and Y. Given $K$ is the kernel of $U^TX$ (e.g., $X^TUU^TX$), L is the kernel of Y (e.g., $Y^TY$), and $H:= I - n^{-1}ee^T$, the problem is posed as a constraint optimization shown below:

\begin{equation}
\begin{aligned}
\underset {U} {argmax}\;\;tr(U^T XHLHX^T U)\\
subject\;to\;U^TU=I
\end{aligned}
\label{equation:Barshan_SPCA}
\end{equation}
Notice the matrix $Q=XHLHX^T$ is real and symmetric, and the eigendecomposition of $Q$ will provide the solution ($U$).

The supervised probabilistic PCA (SPPCA) \citep{yu2006supervised} is a generative model that uses latent variables to generate the original data and class label using EM learning. The generation of observed data ($ x \in  R^M$) and labels ($ y \in  R^L$) from the latent variables ($ z \in R^K$) takes the following form,

\begin{equation}
\begin{aligned}
x = W_x z + \mu_x + \epsilon_x \\
y = W_y z + \epsilon_y
\end{aligned}
\label{equation:SPPCA}
\end{equation}
where $ W_x \in  R^{M \times K}$, $ W_y \in  R^{L \times K}$ are the linear coefficient matrices for data and labels respectively, $ {\mu}_x$ is the data mean, and $ {\epsilon}_x, {\epsilon}_y $ are isotropic Gaussian noise model for ${x},{y}$ respectively. The latent variables are also assumed to follow a Gaussian distribution with zero mean and unit variance and are shared by both inputs and outputs. To generate the class labels, the model uses $L$, where $L$ is the dimension of outcomes, deterministic functions that are assumed to be linear. The learning happens in two steps; in the first step, the expected distribution of $ z$ is calculated while fixing the model parameters. In the next step, the log-likelihood of the data is maximized by keeping the distribution of $ z$ unchanged. These two steps are repeated until the model reaches a local minimum.

Li et al. formulated an SVD (singular value decomposition) based supervised PCA, which they call SupSVD \citep{li2016supervised}. The technique recovers a low-rank approximation of the data matrix $X$ with the help of a supervision matrix $Y$ as shown below

\begin{equation}
\begin{aligned}
 X =  U  V^T +  E, \\
 U =  Y  B +  F
\end{aligned}
\label{equation:SupSVD}
\end{equation}
where $X \in R^{p \times n}$, $Y \in R^{q \times n}$, $U \in R^{n \times r}$  the latent score matrix, $V \in R^{p \times r}$  the full-rank loading matrix, $B \in R^{q \times r}$  the coefficient matrix, and $E \in R^{n \times p}$, $F \in R^{n \times r}$ are two error matrices. The first part of Equation \ref{equation:SupSVD} extracts a low-rank approximation of data matrix $X$, and the second part uses multivariate linear regression to impose the supervision effect of $Y$ on $U$. 

Ritchie et al. proposed another supervised PCA~\citep{ritchie2019supervised}, which minimizes the traditional PCA cost along with a regression loss on the class labels.  Their optimization problem is then 
\begin{equation}
\begin{aligned}
\underset {L, \beta} {minimize}\;\;\|Y -XL^T\beta \|_F^2 + \lambda \|X -XL^TL \|_F^2 \\
subject\;to\;{L}{L}^T={I}_k
\end{aligned}
\label{equation:Richie_LSPCA}
\end{equation}
where $X \in {R^{n\times p}}$ and $Y \in {R^{n\times q}}$ are the data matrix and label respectively, $L \in {R}^{k \times p}$ is the basis for learned subspace, and $\beta \in {R}^{k \times q}$ is the learned coefficients for prediction. The first term is the conventional regression loss calculated on the reduced space ($XL^T$) and the regression coefficient ($\beta$) is the standard least square solution for a fixed $L$. The authors used a gradient-based iterative method to solve the problem over the Grassmannian manifold. Note that the approach doesn't offer a closed-form solution, and the number of principal components is also user-defined.

In contrast to the above-mentioned supervised PCA technique, which can be applied to discrete data points, supervised functional principal component analysis (SFPCA), proposed by Nie et al. \citep{nie2018supervised}, works on functional data analysis. The method finds the functional principal components (FPCs) by maximizing the quantity

\begin{equation}
\begin{aligned}
Q(\xi) = \frac{\theta \langle\xi,\mathcal {L} \xi\rangle + (1-\theta){cov}^2({Y},\langle X,\xi\rangle)}{\| \xi \|^2}\\
subject \;\;to\;\; \langle \xi_i,\xi_i \rangle = 1,\;\; \langle \xi_i,\xi_j \rangle = 0
\end{aligned}
\label{equation:SFPCA}
\end{equation}
where $ X, Y$ are data and labels respectively, $\xi$ the functional principal component, $ 0 \le \theta \le 1 $, $\mathcal {L}$ is the empirical covariance operator, $\langle \cdot,\cdot \rangle$ is the usual inner product space. Notice that the first term in the numerator is the unsupervised FPCA and the second term captures the squared covariance between FPC scores $\langle X,\xi\rangle$ and the response variable ${Y}$. The hyper-parameter $\theta$ balance the unsupervised and supervised terms.

\section{Supervised Linear Centroid-Encoder (SLCE)}
\label{LCE}
Let $X \in \mathbb{R}^{d \times n}$ be a data matrix where $n$ is the total number of samples and $d$ is the 
dimension of each sample $x_i \in \mathbb{R}^d$. Assume the columns of $X$ each belongs to one of $M$ classes $\{C_j\}^M_{ j = 1}$
where the set of pattern indices of class $C_j$ is denoted by $I_j$.  The centroid of each class is defined as
\begin{equation}
c_j=\frac{1}{|C_j|}\sum_{i \in I_j} x_i
\end{equation}
where $|C_j|$ is the cardinality of  class $C_j$.  Define a matrix 
of class means $\tilde{C} \in \mathbb{R}^{d \times n}$ 
where the $i$'th column of $\tilde{C}$ is the centroid associated
with the class of the $i$'th column of $X$.
Note $\tilde{C}$ will have non-unique entries as long as $M<n$. For example, consider the data set $X=\{x_1,x_2,x_3,x_4,x_5\}$ which has two classes $C_1,C_2$ where $I_1=\{1,3,5\}$ 
and $I_2=\{2,4\}$.  Taking $c_1,c_2$ as the corresponding centroids
we have $\tilde C = \{c_1,c_2,c_1,c_2,c_1\}$. With this set up, we present the formulation of Supervised Linear Centroid-Encoder (SLCE).

\subsection{Formulation with Orthogonality Constraint}
\label{formulation_orthogonal_LCE}
The goal of SLCE is to provide the orthogonal projection of the data
to $k$ dimensions that best approximates the class centroids.  
This derivation proceeds one dimension at a time and can be shown
to be equivalent to computing the optimal rank $k$ projection.
 The projection onto the best one-dimensional space spanned by the unknown vector $\textbf{a} \in \mathbb{R}^d$ may be determined by the
optimization problem
\begin{equation}
\begin{aligned}
\underset {{a}} {minimize}\;\;\|\tilde C-{a} {a}^T X \|_F^2 \;\;\;
subject\;to\;{a}^T{a}=1
\end{aligned}
\label{equation:LCE_cost}
\end{equation}

The Lagrangian of Equation (\ref{equation:LCE_cost}) is
\begin{equation}
\begin{aligned}
\mathcal{L}({a},\lambda) = \|\tilde C-{a} {a}^T X \|_F^2 -\lambda ({a}^T{a}-1)
\end{aligned}
\label{equation:LCE_eq7}
\end{equation}
where $\lambda$ is the Lagrangian multiplier. Notice that, setting $\frac{\partial \mathcal{L}}{\partial \lambda}=0$ implies ${a}^T{a} = 1$. Taking the derivative of Equation \ref{equation:LCE_eq7} w.r.t. ${a}$ and setting it to 0 gives,
\begin{equation}
\begin{aligned}
( X \tilde C^T + \tilde C X^T - XX^T) {a} = (\lambda + {a}^T X X^T {a}){a}
\end{aligned}
\label{equation:LCE_eq10}
\end{equation}
Setting $\mu = \lambda + {a}^T X X^T {a}$ we obtain
\begin{equation}
\begin{aligned}
(X \tilde C^T + \tilde C X^T - XX^T) {a} = \mu {a}
\end{aligned}
\label{equation:LCE_eq11}
\end{equation}
Since $ X \tilde C^T + \tilde C X^T -XX^T$ is a  symmetric matrix, 
the solution to the optimization problem given by Equation
(\ref{equation:LCE_eq7}) is provided by 
the eigenvector of Equation (\ref{equation:LCE_eq11})
with the largest eigenvalue.
To solve for the second projection direction ${b}$ we require
\begin{equation}
\begin{aligned}
\underset {{b}} {minimize}\;\;\|\tilde C-{b} {b}^T X \|_F^2\\
subject\;to\;{b}^T{b}=1 \;\;\;{b}^T{a}=0
\end{aligned}
\label{equation:centroid_reconstruction_PCA_2ndvec}
\end{equation}
The Lagrangian of Equation \ref{equation:centroid_reconstruction_PCA_2ndvec}  is
\begin{equation}
\begin{aligned}
\mathcal{L}(b, \alpha, \beta) = \|\tilde C-{b} {b}^T X \|_F^2 -\alpha ({b}^T{b}-1) - \beta ({b}^T{a})
\end{aligned}
\label{equation:centroid_reconstruction_PCA_2ndvec_eq1}
\end{equation}
where $\alpha$ and $\beta$ are the Lagrangian multipliers. 
Differentiating the Lagrangian again and setting equal to zero we
obtain the necessary condition 
for ${b}$
\begin{equation}
\begin{aligned}
(X \tilde C^T + \tilde C X^T - XX^T) {b} =  \gamma {b}\\
\
where\;\;\;\gamma = (\alpha + {b}^T XX^T {b})
\end{aligned}
\label{equation:centroid_reconstruction_PCA_2ndvec_eq5}
\end{equation}
Here ${b}$ is the eigenvector of the symmetric matrix $X \tilde C^T + \tilde C X^T - XX^T$ associated with the second
largest eigenvalue and (from symmetry) ${b}$ is orthogonal to ${a}$.
One can proceed in a similar fashion 
and obtain the remaining 
solutions~\citep{horn2012matrix}.

\subsection{Formulation as a  Rank-$k$ Projection}

To find the best rank-$k$ projection we can solve the optimization problem
\begin{equation}
\begin{aligned}
\underset {{A}} {minimize}\;\;\|\tilde C-{A} {A}^T X \|_F^2 \;\;\;
subject\;to\;{A}^T{A} = I \\
where\;\; A\in \mathbb{R}^{d\times k}
\end{aligned}
\label{equation:LCE_cost_k-dim}
\end{equation}
The Lagrangian of Equation (\ref{equation:LCE_cost_k-dim}) is 
\begin{equation}
\begin{aligned}
\mathcal{L}({A},\Lambda) = \|\tilde C-{A} {A}^T X \|_F^2 -tr(\Lambda ({A}^T{A}-I))
\end{aligned}
\label{equation:LCE_cost_k-dim_equ1}
\end{equation}
where $\Lambda \in \mathbb{R}^{k \times k}$ is a symmetric matrix which contains the Lagrangian multipliers. It can be shown that $\Lambda$ will be an diagonal matrix where the diagonal entries are the Lagrange multipliers \footnote{We have left the proof in Appendix \ref{diag_lambda}}. Differentiating the Equation (\ref{equation:LCE_cost_k-dim_equ1}) w.r.t. $A$ and setting the derivative to zero gives

\begin{equation}
\begin{aligned}
( X \tilde C^T + \tilde C X^T - XX^T) {A} = {A} \Lambda
\end{aligned}
\label{equation:LCE_cost_k-dim_eq2}
\end{equation}

\begin{figure}[!ht]
    \centering
    \includegraphics[width=15.00cm,height=7.5cm]{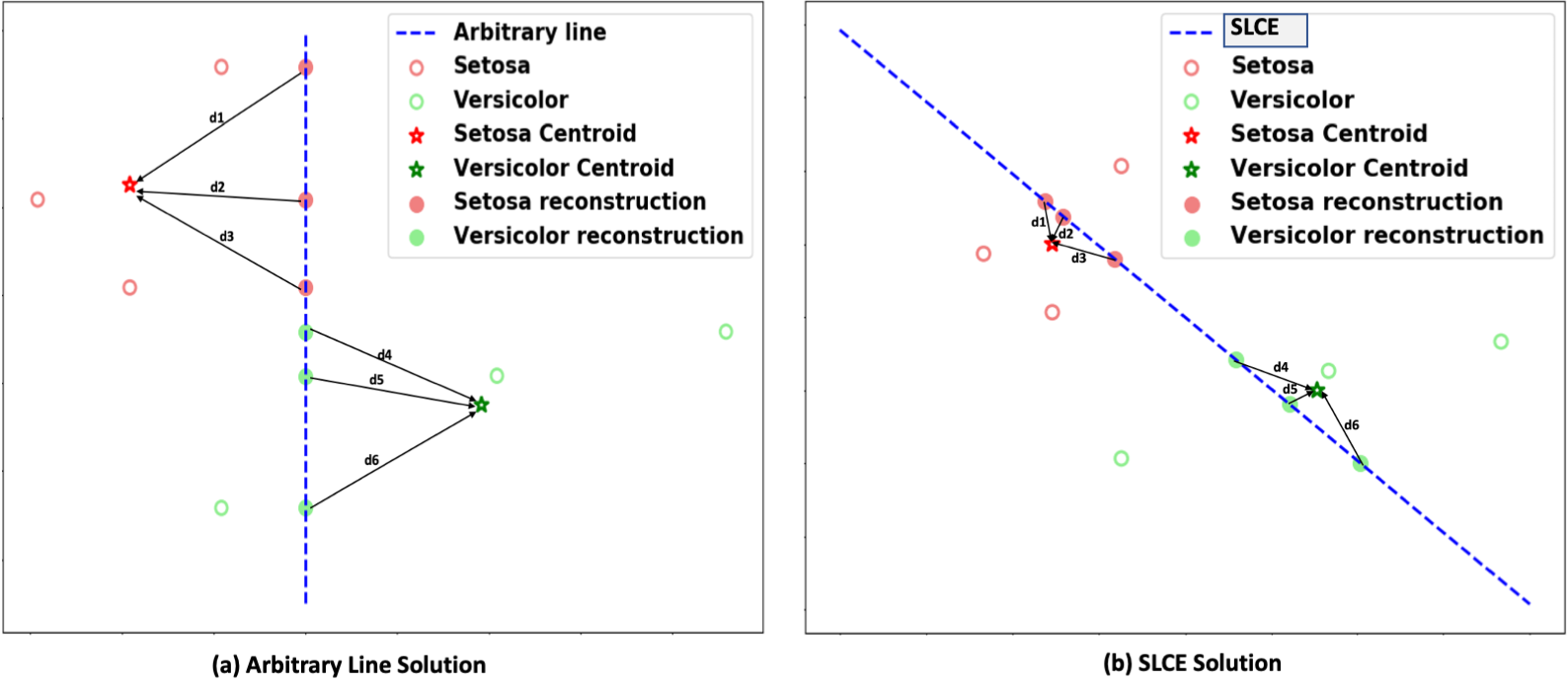}
    \vspace{-0.25cm}
    \caption{The geometric intuition of the SLCE algorithm using Setosa and Versicolor classes of Iris data, where we used the first two features to represent each sample. In panel (a), we used an arbitrary line to compute the cost in Equation~(\ref{equation:LCE_cost}). The original samples were reconstructed using the line, and then the distances from the corresponding class centroid were denoted using the black lines. On the other hand, panel (b) shows the reconstruction of all the samples using the SLCE solution and the distances from the corresponding class centroids. Notice that the sum of the distances ($d_1,...,d_6 $) using SLCE is less than using an arbitrary line; SLCE explicitly searches for a line that minimizes the sum of the distances.}
    \label{fig:LCE_Visual}
\end{figure}

In Figure (\ref{fig:LCE_Visual}) we illustrate the geometric intuition behind SLCE.  This figure captures how SLCE produces a subspace to
to reduce the data that captures label information through the centroids.

\subsection{Some properties of SLCE}
Here we present several of mathematical properties of the proposed algorithm.


\medskip

\textbf{Property 1. The matrix $\mathbf{ X \tilde C^T + \tilde C X^T}$ is a symmetric and positive semi-definite (PSD).}

Proof: By construction $\tilde C$ has the corresponding $c_j$'s for each $x_i$ in $X$ and both $\tilde C,X \in \mathbb{R}^{d \times n}$. Without loss of generality, we can order the samples in $X$ based on the corresponding class label, i.e., all the samples of class $C_1$ will appear first, followed the samples of class $C_2,C_3,...,C_M$ where $M$ is the total number of classes. Lets consider the representation of the term $X \tilde C^T$, i.e., 
\begin{equation}
\begin{aligned}
 X \tilde C^T = \sum_{j=1}^{M}\sum_{i \in I_j} x_i c_j^T
\end{aligned}
\label{equation:psd_eq1}
\end{equation}
where $I_j$ is the index set of $j^{th}$ class.  This can be rewritten as
\begin{equation}
\begin{aligned}
 X \tilde C^T = \sum_{j=1}^{M} \left (\sum_{i \in I_j} x_i \right ) c_j^T
\end{aligned}
\label{equation:psd_eq2}
\end{equation}
from which it follows
\begin{equation}
\begin{aligned}
 X \tilde C^T = \sum_{j=1}^{M}|C_j| \left ( \frac{1}{|C_j|}\sum_{i \in I_j} x_i \right )c_j^T
\end{aligned}
\label{equation:psd_eq3}
\end{equation}
Defining
$|C_j|$ as the cardinality of $j^{th}$ class we see
\begin{equation}
\begin{aligned}
 X \tilde C^T = \sum_{j=1}^{M}|C_j| \left(c_j c_j^T\right )
\end{aligned}
\label{equation:psd_eq4}
\end{equation}

Notice, each $c_j c_j^T$ is PSD. As $X \tilde C$ is a sum of $M$ PSD matrices, hence $X \tilde C$ is a symmetric PSD matrix. 
Hence 
\begin{equation}
\tilde C X^T = X\tilde C^T
\end{equation}
so it 
is also PSD. Therefore $X \tilde C^T + \tilde C X^T$ is a symmetric PSD matrix.

\medskip
Given the construction from above, we have the following additional property:

\textbf{Property 2. If $\mathbf{M}$ is the number of classes, $\mathbf{X}$ and $\mathbf{\tilde{C}}$ the data and centroid matrix, respectively,   then the 
rank $(\mathbf{X \tilde C^T + \tilde C X^T}) \mathbf{\le M-1}$.}

Proof: 
From Property 1. we know
rank $(X \tilde C^T + \tilde C X^T) = rank(\tilde C X^T)$.  We also know that $rank( \tilde C X^T ) = rank( \tilde C)$.
Note, mean subtraction of data matrix $X$ makes the class centroid as linearly dependent\footnote{See Appendix \ref{Liner_dependency}}. Hence rank$(X \tilde C^T + \tilde C X^T)  = rank( \tilde C )\le (M-1)$.

This next property is important in that it tells us which eigenvectors to use for the
data reduction.

\textbf{Property 3. The matrix $ \mathbf{ X \tilde C^T + \tilde C X^T - XX^T}$ has at most $\mathbf{M-1}$ positive eigenvalues where $\mathbf{M}$ is the number  of classes in $\mathbf{\tilde{C}}$.}

\begin{figure}[!ht]
    \centering
    \includegraphics[width=12.0cm,height=8.25cm]{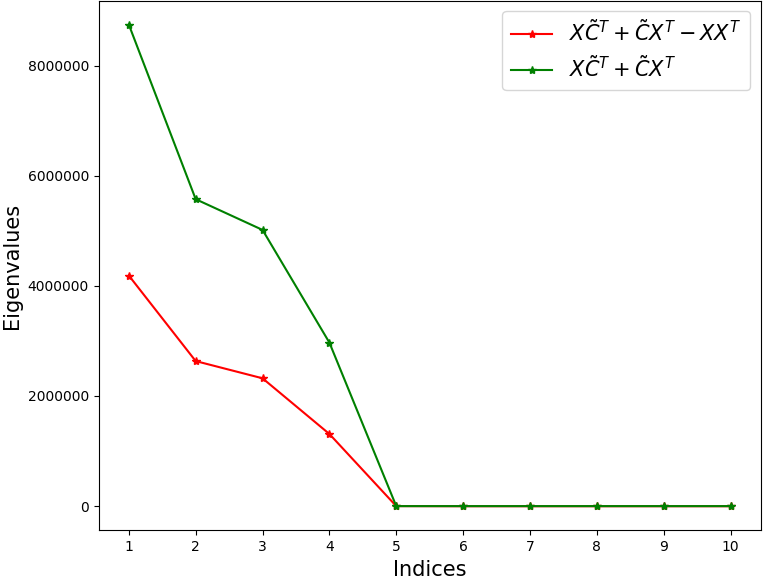}
    \vspace{-0.25cm}
    \caption{ Plot of the first ten eigenvalues of $(X \tilde C^T + \tilde C X^T - XX^T)$ and $(X \tilde C^T + \tilde C X^T)$ using PANCAN data set which has five classes. The first four eigenvalues of $(X \tilde C^T + \tilde C X^T - XX^T)$ are positive and bounded above by the eigenvalues of $(X \tilde C^T + \tilde C X^T)$.}
    \label{fig:LCE_theorem}
\end{figure}

Proof: Let $A := X \tilde C^T + \tilde C X^T - XX^T $ and $B := XX^T$ where $A,B \in \mathbb{R}^{d \times d}$. $A+B=X \tilde C^T + \tilde C X^T$. 
From the Monotonicity theorem, which is a corollary of Weyl's theorem involving inequalities of Hermitian matrices \citep{horn2013matrix} we have
\begin{equation}
\begin{aligned}
\lambda_i (A+B) \ge \lambda_i (A) 
\end{aligned}
\label{equation:positive_eigenvalue_eq1}
\end{equation}
where we assume the eigenvalues are in decreasing order.
Hence we conclude
\begin{equation}
\begin{aligned}
\lambda_i (X \tilde C^T + \tilde C X^T) \ge \lambda_i (X \tilde C^T + \tilde C X^T -XX^T) 
\end{aligned}
\label{equation:positive_eigenvalue_eq2}
\end{equation}
 Since $\lambda_i(X \tilde C^T + \tilde C X^T) = 0$ if $i\ge M$, it follows that 
$$\lambda_i (X \tilde C^T + \tilde C X^T -XX^T) \le 0$$
for $i \ge M$ from which the property follows.
The Figure (\ref{fig:LCE_theorem}) demonstrates the inequality on PANCAN data.


There is also an important relationship 
between the eigenvalues and the objective function.

\textbf{Property 4. Let $\mathbf{a}$ be an eigenvector of the matrix
$\mathbf{X \tilde C^T + \tilde C X^T - XX^T}$ with eigenvalue $\mu$.  Then it follows
$$\mathbf{\|\tilde C-{a} {a}^T X \|^2_F = Tr(\tilde C^T \tilde C) - \mu}$$}

\begin{figure}[!ht]
    \centering
    \includegraphics[width=12.0cm,height=8.25cm]{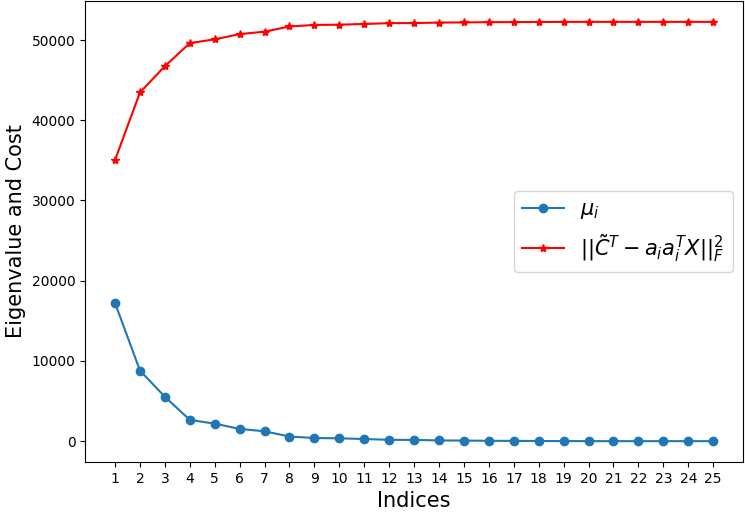}
    \vspace{-0.25cm}
    \caption{We show how the one-dimensional reconstruction cost (Equation \ref{equation:LCE_cost}) changes for the first 25 eigenvector along with the corresponding eigenvalues. The experiment is run using COIL20 dataset.}
    \label{fig:LCE_Recons_Eigenval}
\end{figure}

Proof:
Given 
$$(X \tilde C^T + \tilde C X^T - XX^T) {a} = \mu a$$
it follows that
\begin{equation}
\begin{aligned}
 {a}^T(X \tilde C^T + \tilde C X^T - XX^T) {a} = \mu
\end{aligned}
\label{equation:centroid_reconstruction_PCA_eq10}
\end{equation}
and, after taking the trace and using
its properties, 
\begin{equation}
\begin{aligned}
 -Tr({a}^T X \tilde C^T {a}) - Tr({a}^T \tilde C X^T {a}) + Tr({a}^T XX^T {a}) = -\mu
\end{aligned}
\label{equation:centroid_reconstruction_PCA_eq12}
\end{equation}
Now adding $Tr(\tilde C^T \tilde C)$ to the both side of Equation \ref{equation:centroid_reconstruction_PCA_eq12}

\begin{equation}
\begin{aligned}
Tr(\tilde C^T \tilde C - \tilde C^T {a}{a}^T X - X^T {a}{a}^T \tilde C + X^T {a}{a}^T {a} {a}^T X) = Tr(\tilde C^T \tilde C) - \mu
\end{aligned}
\label{equation:centroid_reconstruction_PCA_eq16}
\end{equation}

\begin{equation}
\begin{aligned}
Tr[(\tilde C-{a} {a}^T X)^T (\tilde C-{a} {a}^T X)] = Tr(\tilde C^T \tilde C) - \mu
\end{aligned}
\label{equation:centroid_reconstruction_PCA_eq17}
\end{equation}

\begin{equation}
\begin{aligned}
\|\tilde C-{a} {a}^T X \|^2_F = Tr(\tilde C^T \tilde C) - \mu
\end{aligned}
\label{equation:centroid_reconstruction_PCA_eq18}
\end{equation}
The above equation establishes the relationship between the cost and the eigenvalue. If $\mu_1 > \mu_2$ and the corresponding costs are $C_{\mu_1},C_{\mu_2}$, then $C_{\mu_1}<C_{\mu_2}$.
Figure \ref{fig:LCE_Recons_Eigenval} demonstrates the relationship visually. 

\textbf{Property 5. Let $\mathbf{A}$ be a rank $\mathbf{k}$ orthonormal matrix
that minimizes the error in Equation
\ref{equation:LCE_cost_k-dim}.   Then the total error in the approximation is}
   \begin{equation}
   \label{error-eq}
   \mathbf{\|\tilde C-{A} {A}^T X \|^2_F = Tr(\tilde C^T \tilde C) - \sum_{i=1}^k \mu_i}
   \end{equation}

Proof:  the derivation is similar to
that of Equation (\ref{equation:centroid_reconstruction_PCA_eq18}).

\textbf{Property 6. The subspace of best fit given by the range of $\mathbf{A}$, $\mathbf{\mathcal{R}(A)}$,  has $k$ dimensions where $\mathbf{k} \le \mathbf{M-1}$ is the number of non-zero eigenvalues $\mu_i$.}

 Proof: This fact follows directly from the error Equation (\ref{error-eq}).  The error is no longer reduced when the eigenvalues become zero.  From Property 3 the number of positive eigenvalues
 is bounded by $M-1$.  If we take additional eigenvectors to represent the data they will not decrease the error in the objective function if the eigenvalue is zero and will increase the error if the eigenvalue is negative.


\newpage
\subsection{Connection with PCA}
PCA can be derived from the reconstruction perspective as follows,

\begin{equation}
\begin{aligned}
\underset {{a}} {minimize}\;\;\|X -{a} {a}^T X \|_F^2\\
subject\;to\;{a}^T{a}=1
\end{aligned}
\label{equation:PCA}
\end{equation}
where ${a}$ is the transformation vector. Comparing equation \ref{equation:PCA} with \ref{equation:LCE_cost} we can say PCA reconstructs each samples, whereas SLCE reconstructs the centroid of a class. As SLCE uses the class labels to calculate the centroids, hence it can be thought of as supervised PCA. 

\begin{figure}[!ht]
    \centering
    \includegraphics[width=15.0cm,height=12.0cm]{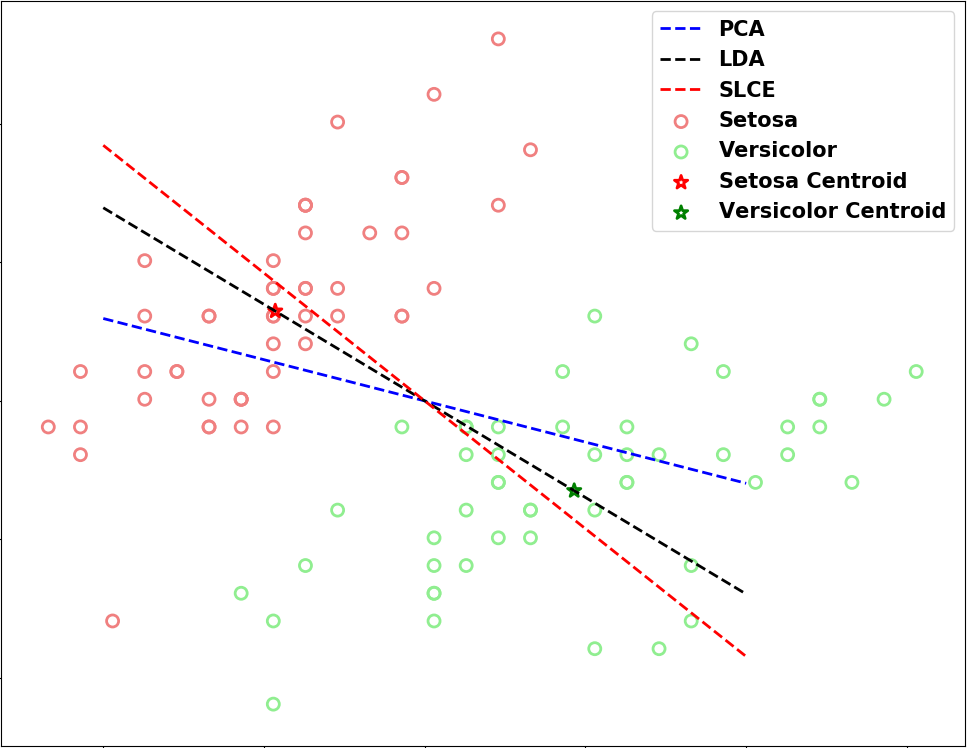}
    \vspace{-0.25cm}
    \caption{Comparison of PCA,LDA, and SLCE solution on Iris data. We used the first two features to represents each samples from Setosa and Versicolor classes.}
    \label{fig:PCA_LCE_Comparison}
\end{figure}
Figure \ref{fig:PCA_LCE_Comparison} compares the solutions of PCA and SLCE on a toy data. Notice that the first eigenvector of PCA is governed by the data variance where as the solution of SLCE is governed by the class centroids.

\begin{figure}[!ht]
    \centering
    \includegraphics[width=15.0cm,height=12.0cm]{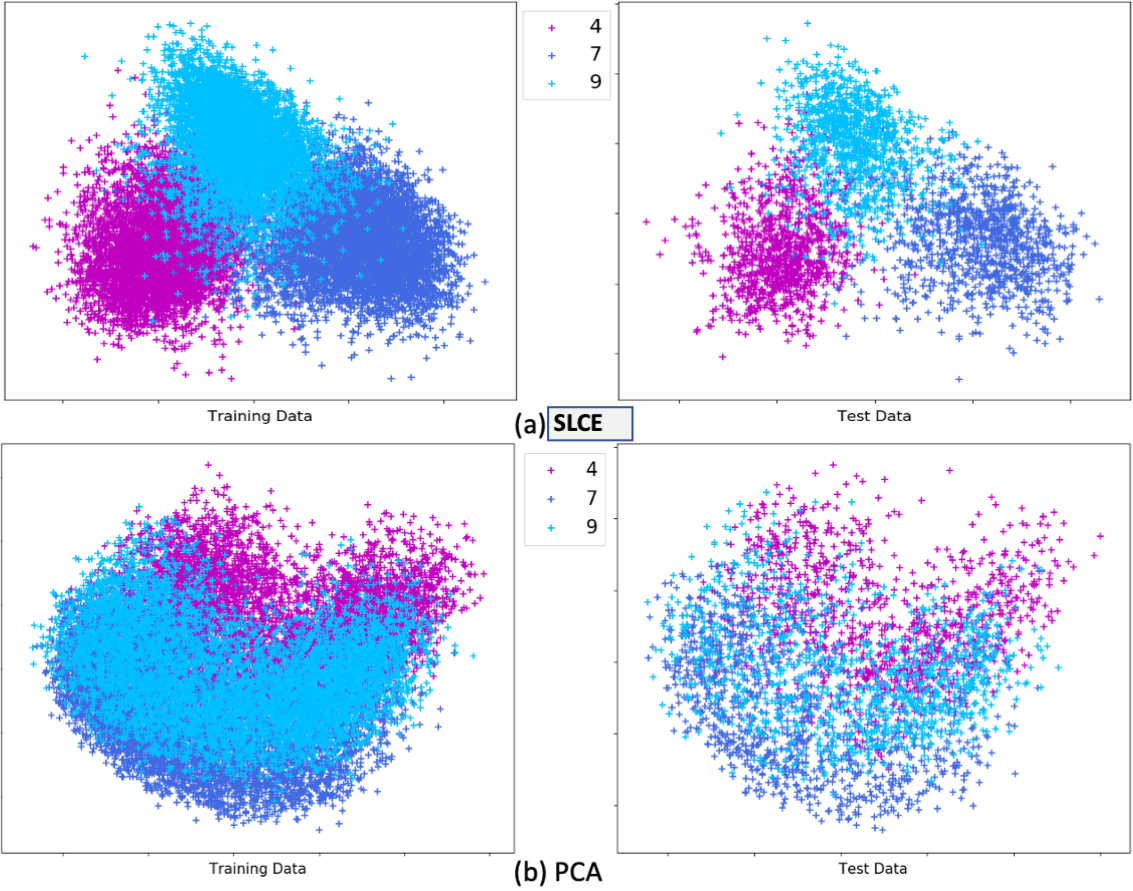}
    \caption{Comparison of SLCE (a) and PCA (b) projection on MNIST digits 4,7, and 9.}
    \label{fig:LCE-vs-PCA_MNIST-4-7-9}
\end{figure}

In Figure \ref{fig:LCE-vs-PCA_MNIST-4-7-9}, we show the visualization on MNIST digits 4, 7, and 9. These three digits are not separable in two-dimensional PCA space, and we want to verify whether SLCE can produce better visualization. In panel (a), we present the 2D SLCE projection. The three test classes are clearly separated, creating three blobs for each category.

\subsection{Connection with Centroid-Encoder}
The nonlinear mapping of centroid-encoder (CE) \citep{ghosh2022supervised} is defined as,
\begin{equation}
\begin{aligned}
 \underset {\theta} {\text{minimize}}\;\; \sum^M_{j=1} \sum_{i \in I_j}\|c_j-f(x^i; \theta))\|^2_2
 \label{equation:CECostFunction1}
\end{aligned}
\end{equation}
where $f(\cdot) = h(g(\cdot))$, composition of dimension reducing {\it encoder}
mapping $g$ followed by a dimension increasing {\it decoder} mapping $h$ with $\theta$ being the parameter set for the mapping function. Both SLCE and CE reconstruct a class-centroid from the samples belonging to that class, but unlike CE, SLCE incorporates a linear mapping with orthogonality constraints.

\section{Visualization and Classification Results}
\label{viz_clf_result}
We present the comparative evaluation of our models on various data sets using several linear dimensionality reduction techniques. We evaluated our proposed approach, SLCE as described in Section\ref{formulation_orthogonal_LCE} to compare and contrast with other five State-of-the-Art techniques. 

\subsection{Experimental Details}
This section compares our proposed method with other linear dimensionality reduction techniques on eleven benchmarking data sets.  Table \ref{table:dataDescription} gives the data set details which were also used in literature \citep{Barshan:2011:SPC:1950989.1951174,ritchie2019supervised}. We ran three sets of experiments to compare our method with Fisher LDA \citep{duda2006pattern}, Bairs Supervised PCA \citep{doi:10.1198/016214505000000628}, Barshan's HSIC Supervised PCA \citep{Barshan:2011:SPC:1950989.1951174}, Richie's Supervised PCA \citep{ritchie2019supervised,ritchie2020supervised},and PCA \citep{jolliffe_1986}. We didn't compare the SupSVD method \citep{li2016supervised} as the authors didn't run any classification experiment in reduced space.
\begin{table}[!ht]	
	\centering
	\begin{tabular} {|c|c|c|c|c|c|}	
		\hline	
		Dataset & \#Features & \#Classes & \#Samples & Domain \\
		\hline
	    USPS & 256 & 10 & 11000 & Image \\
        \hline
        MNIST & 784 & 10 & 70000 & Image \\
        \hline
        \multirow{2}{*} {Human Activity} & \multirow{2}{*} {561} & \multirow{2}{*} {6} & \multirow{2}{*} {5744} & Accelerometer \\
        & & & & Sensor\\
        \hline
        Ionosphere & 34 & 2 & 354 & Radar  \\
        \hline
        Colon & 2000 & 2 & 62 & Biology \\
        \hline
        Mice Protein & 77 & 8 & 975 & Biology \\
        \hline
		\multirow{2}{*}{Arcene} & \multirow{2}{*}{10000} & \multirow{2}{*}{2} & \multirow{2}{*}{900}  & Mass \\
        & & & & Spectrometric \\
        \hline
        PANCAN & 20531 & 5 & 801 &  RNA-Seq \\
        \hline
        Olivetti & 4096 & 20 & 400 & Image \\
        \hline
        Yale Face & 1024 & 15 & 165 & Image \\
        \hline
        COIL20 & 1024 & 20 & 1440 & Image \\
		\hline
	\end{tabular}	
	\caption{Descriptions of the data sets used for bench-marking experiments.}
	\label{table:dataDescription}
\end{table}
The experiments follow the standard workflow.
\begin{itemize}
    \item  \textbf{Step1:} Split each data sets into training and test partition.
    \item \textbf{Step2:} Train each models on the training set. 
    \item \textbf{Step3:} Using the trained models, project the training and test samples on $p$-dimensional space where $p \in \{2,3,5,10,15,20\}$.
    \item \textbf{Step4:} Calculate $5-$NN accuracy on the $p-$dimensional space.
    \item \textbf{Step5:} Repeat steps 1 to 4 twenty-five times and report average accuracy with standard deviation.
\end{itemize}
The experiment-specific details are given below.\\
\textbf{Experiment 1:}
This experiment aims to compare each model's low-dimensional ($p \in \{2,3\}$) embedding on USPS, MNIST, Human Activity, Mice Protein,  Arcene, and PanCan data sets. The comparison is made using a $5-$NN classifier on reduced space. We split each data set into a ratio of 80:20 of training and test partition, except for MNIST, which has a separate test set. We repeat the process 25 times and report the average accuracy with standard deviation.\\
\textbf{Experiment 2:} In this experiment, we compared SLCE with Richie's Supervised PCA \citep{ritchie2020supervised} on Colon, Ionosphere, and Arcene data sets. Following the experimental setup in \citep{ritchie2020supervised}, we split each dataset into an $80:20$ ratio of train and test. We built our models on the training set to embed data on two-dimensional space; we then predicted the class of test samples using a $5-$NN classifier. We compared our method with the published results in \citep{ritchie2020supervised}.\\
\textbf{Experiment 3:} The goal of this experiment is to compare the models by performing classification on different embedding dimensions, i.e., $p \in \{5,10,15,20\}$. We used three data sets, Olivetti, YaleFace, and COIL20, and split them into a $50:50$ ratio of train and test. Each model is fitted on the training partition, and the $5-$NN classification is calculated using different embedding dimensions. The process is repeated 25 times, and the average accuracies are plotted for comparison.

\subsection{Results}
\begin{figure}[!ht]
    \includegraphics[width=15.0cm,height=7.0cm]{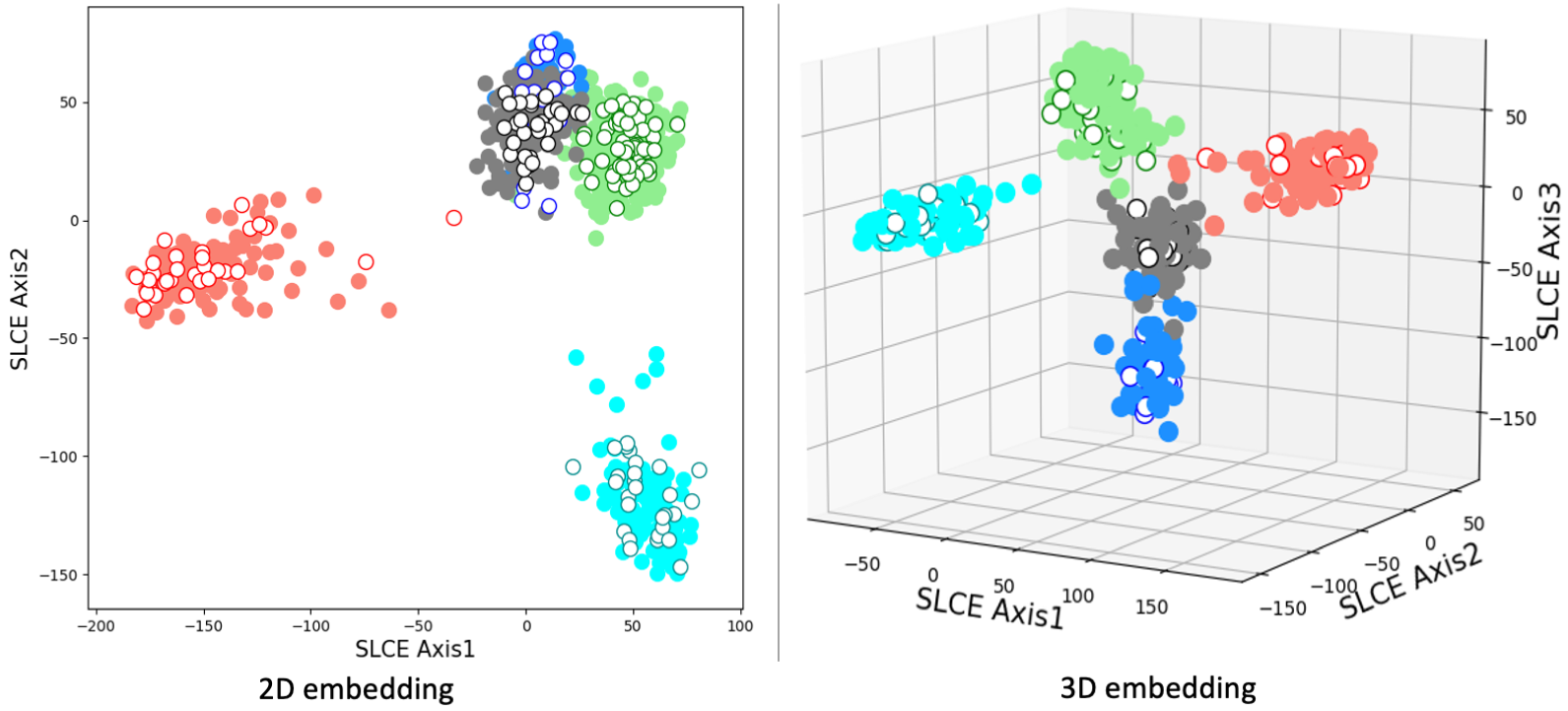}
    \caption{Embedding of PANCAN data in two (left) and three (right) using SLCE. The solid and blank circles are the training and test cases respectively.}
    \label{fig:LCE_PanCan_2D_3D}
\end{figure}

First, we discuss the results of \textit{Experiment 1} comparing SLCE, LDA, PCA, HSIC PCA, and Bair's SPCA as presented in Table~\ref{table:LCE_PCA_SPCA_comparison}. We used embedding dimensions 2 and 3. 
We observe that SLCE generally produces better generalization performance than other methods, both in the two and three-dimensional embedding space. 
\begin{table}[!ht]
	\begin{tabular} {|c|c|c|c|c|c|}
 	\hline\hline 
	\multicolumn{6}{|c|} {Classification on embedding dimension = 2}\\
  \hline
  \multirow{1}{*}{Dataset} & \multicolumn{1}{c|} {SLCE} & \multicolumn{1}{c|} {LDA} & \multicolumn{1}{c|} {PCA} & \multicolumn{1}{c|} {HSIC SPCA}& \multicolumn{1}{c|} {Bair's SPCA}\\
		
		\hline
		USPS & 55.64 $\pm$ 0.95 & $47.20 \pm 0.77$ & $39.83 \pm 0.91$ &  $45.67 \pm 1.16$ & $40.88 \pm 0.92$\\
		\hline
		MNIST & $45.74 \pm 0.00$ & $46.03 \pm 0.00$ & $42.43 \pm 0.00$ & $44.17 \pm 0.00$ & $41.92 \pm 0.00$\\
		\hline
        Activity & 65.20 $\pm$ 0.87 & $64.18 \pm 1.36$ & $51.83 \pm 1.46$ & $63.26 \pm 1.26$ & $53.05 \pm 1.76$\\
        \hline
        Mice Protein & $65.56 \pm 2.83$ & $51.50 \pm 2.37$& $44.91 \pm 2.82$ & $51.54 \pm 3.59$ & $64.57 \pm 9.24$\\
        \hline
        Arcene & $83.61 \pm 5.05$ & $63.71 \pm 5.70$ & $67.71 \pm 7.63$ & $71.80 \pm 6.25$ & $68.98 \pm 5.14$\\
        \hline
        PANCAN & $88.42 \pm 2.44$ & $94.23 \pm 4.78$ & $94.63 \pm 1.71$ & $94.65 \pm 1.95$ & $94.55 \pm 2.27$\\
		\hline \hline
    \multicolumn{6}{|c|} {Classification on embedding dimension = 3}\\
        \hline
    	USPS & \textbf{76.69} $\pm$ \textbf{0.69} & $69.37 \pm 1.15$ & $47.12 \pm 0.78$ &  $66.99 \pm 0.81$ & $48.79 \pm 1.80$\\
		\hline
		MNIST & \textbf{69.72} $\pm$ \textbf{0.00} & $67.51 \pm 0.00$ & $48.75 \pm 0.00$ & $63.59 \pm 0.00$ & $49.90 \pm 0.00$\\
		\hline
        Activity & \textbf{77.98} $\pm$ \textbf{1.21} & $70.57 \pm 1.02$ & $66.96 \pm 1.04$ & $70.87 \pm 1.16$ & $66.91 \pm 1.13$\\
        \hline
        Mice Protein & \textbf{80.56} $\pm$ \textbf{2.94} & $68.50 \pm 3.32$& $64.46 \pm 3.11$ & $67.89 \pm 2.70$ & $77.09 \pm 5.05$\\
        \hline
        Arcene & $\textbf{84.00} \pm \textbf{6.62}$ & $62.54 \pm 8.70$ & $71.22 \pm 6.90$ & $76.68 \pm 5.82$ & $69.07 \pm 6.89$\\
        \hline
        PANCAN & $98.94 \pm 0.89$ & $\textbf{99.09} \pm \textbf{0.85}$ & $98.60 \pm 0.73$ & $94.53 \pm 1.12$ & $98.28 \pm 1.01$\\
		\hline \hline
	\end{tabular}
	\caption{Classification accuracies ($\%$) of $5$-NN classifier on the 2D and 3D embedded data by various dimensionality reduction techniques. The best results are highlighted in bold.}	

	\label{table:LCE_PCA_SPCA_comparison}
\end{table}
Notice that  PCA performed better than LDA on the Arcene data. Because Arcene has two classes, the LDA classification occurs in 1-dimensional space, whereas PCA uses two-dimensional space. We think the extra dimension of PCA makes it better than LDA in this case. Observe that SLCE performed poorly on PANCAN data in 2D space, but the performance improved significantly in 3D. By design SLCE creates the embedding by reconstructing the class centroids in the ambient dimension. If centroids are close to each other in the ambient space, SLCE will put the classes close together in the low-dimensional space deteriorating the classification. But in a higher dimension, the classes could be separated, improving the classification result. The visualization in Figure \ref{fig:LCE_PanCan_2D_3D} of the PANCAN data using SLCE establishes the fact. Notice that the samples from black and blue categories are on top of each other in 2D space but are well separated in 3D space, which clarifies the jump in classification accuracy.

\begin{figure}[!ht]
    \centering
    \includegraphics[width=12.0cm,height=8.5cm]{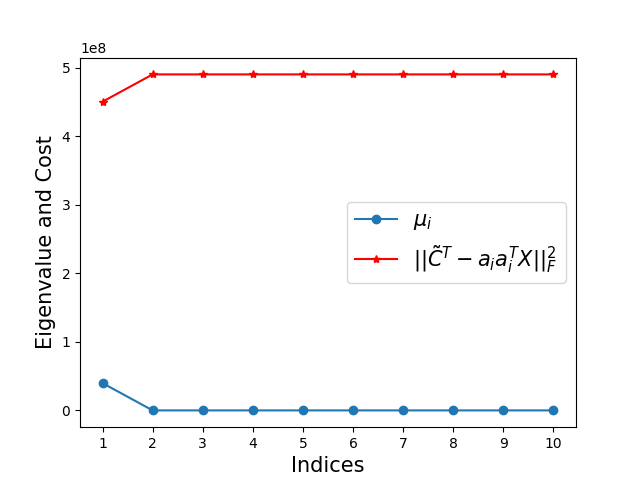}
    \caption{Eigenvalue ($\mu_i$) and the corresponding centroid reconstruction loss ($||\tilde C^T - a_i a_i^T X||_F^2$) of the first ten eigenvectors of Arcene data.}
    \label{fig:LCE_Cost_Eigenvalues_Arcene}
\end{figure}

The classification of 
the Arcene dataset in 3D space has mostly stayed the same compared to 2D for SLCE. As Arcene has two classes, the first eigenvector of SLCE has a positive eigenvalue, and the following two solutions have zero eigenvalues, see Figure \ref{fig:LCE_Cost_Eigenvalues_Arcene}. Therefore the second and third eigenvectors don't minimize the centroid reconstruction loss much compared to the first one; hence they don't contribute much to classification. In fact, classification accuracy using the first eigenvector is $83.12 \pm 6.28$, which is as good as using two/three dimensions. Not surprisingly, in most cases, supervised methods perform better than PCA, which doesn't use labels. The standard deviation on MNIST is 0 for all the models given MNIST has a fixed training and test partition.

Table \ref{table:exp2_results} gives quantitative measures of the quality of embedding comparing SLCE with LRPCAs. 
\begin{table}[ht!]
	\centering
	\begin{tabular} {|c|c|c|c|}	
		\hline
		\multirow{2}{*}{Data set} & \multicolumn{3}{c|} {Supervised Methods} \\
		\cline{2-4}
		& \multicolumn{1}{c|} {SLCE} & \multicolumn{1}{c|} {LRPCA (CV)} &  \multicolumn{1}{c|} {LRPCA (MLE)}\\
		\hline
		Colon & $\textbf{83.08} \pm \textbf{7.22}$ & $80.80 \pm 10.40$ & $80.80 \pm 12.50$ \\
		\hline
        Ionosphere & $\textbf{86.03} \pm \textbf{4.36}$ & $83.90 \pm 4.20$ & $85.90 \pm 2.60$ \\
        \hline
        Arcene & $\textbf{83.41} \pm \textbf{6.51}$ & $80.67 \pm 11.20$ & $81.00 \pm 8.40$ \\
        \hline
	\end{tabular}	
	\caption{Comparison of classification results between SLCE and LRPCA on several bench-marking data sets. The mean classification accuracies are measured in two-dimensional space over 25 runs. Results of LRPCA are reported from \citep{ritchie2020supervised}. The best result is highlighted in bold.}
	\label{table:exp2_results}
\end{table}
SLCE stood out as the best-performing model in all three scenarios. Notice that the standard deviation is also better in all the cases except for Ionosphere. 

\begin{figure}[!ht]
    \centering
    \vspace{-1.0cm}
    \includegraphics[width=12.0cm,height=17.0cm]{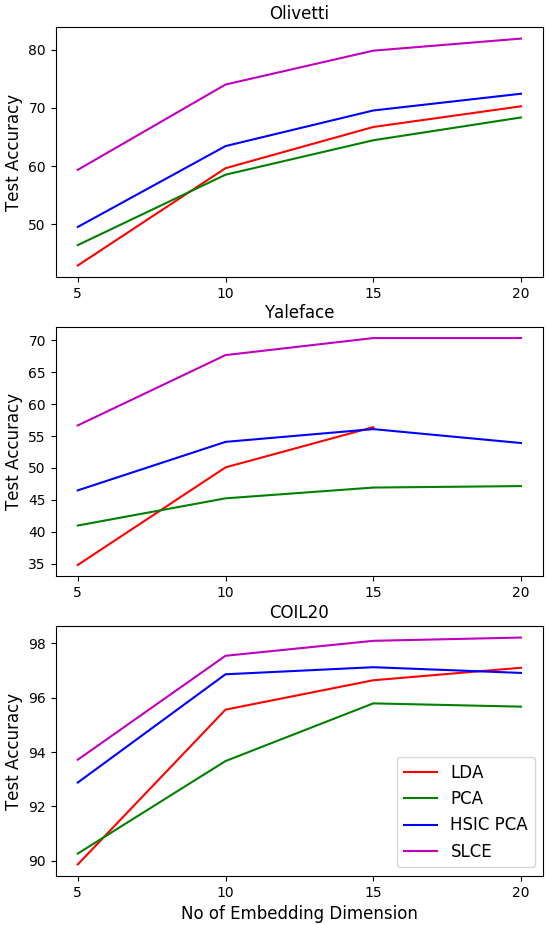}
    \caption{Comparison of classification accuracies on Olivetti (top), Yaleface (middle), and COIL20 (bottom) data using LDA, PCA, HSIC PCA, and SLCE. For each data set, classification is done on four different embedding dimensions, 5, 10, 15, and 20. As Yale Face has 15 classes so the maximum embedding dimension for LDA is 14 ($\#class-1$).}
    \label{fig:PCA_LCE_HSIC_LDA_Comparison}
\end{figure}
Now we turn our attention to the third experiment, where the goal is to plot the classification accuracies as a function of embedding dimension. Figure \ref{fig:PCA_LCE_HSIC_LDA_Comparison} presents the classification accuracies using different embedding dimensions on Olivetti, Yaleface, and COIL20 data. In general, the accuracy increases with the embedding dimension across the methods. Our proposed method SLCE significantly outperform other supervised models in all three cases. The Yaleface has 15 classes, so the LDA classification uses 14-dimensional space. As in Experiment 1, PCA performed poorly in most classification tasks except for three cases with embedding dimension five, where PCA showed better performance than LDA. Note SLCE didn't improve the accuracy on Yaleface from embedding dimensions 15 to 20. The reason is the same as in Arcene. The top fourteen eigenvectors have positive eigenvalues, which minimizes the cost, and any eigenvector after that doesn't reduce the cost; hence the classification accuracy remains the same.

\newpage
\section{Conclusion and Future Work}
\label{conc_future_work}
We proposed a supervised dimensionality reduction technique SLCE. We formulated a constrained optimization problem and showed that it has a closed-form solution using an eigendecomposition.
We proved the eigenvalue problem has at most $M-1$ number of positive eigenvalues where $M$ is the number of classes. We established a connection between the eigenvalues and the cost of the model. We further demonstrated that our proposed model, SLCE is a form of supervised PCA. Unlike other supervised PCA formulations, our model doesn't use raw class labels; instead, it uses the class centroid to impose the supervision. At the same time, the model doesn't require tuning any hyperparameters using a validation set as needed by Bair's SPCA and Richie's SPCA. 
Unlike Centroid-Encoder, SLCE doesn't require tuning a neural network architecture and other hyperparameters, e.g., learning rate, minibatch size, etc.
The closed form solution is appealing and SLCE can be used for much smaller datasets than required by nonlinear data fitting problems.

Our classification and visualization experiment on various data sets established the efficiency of the proposed method and showed that, in most cases, it produces better generalization performance compared to other linear State-of-the-Art techniques. The computational complexity of SLCE is the same as PCA, with an added overhead to compute the class centroids, but this is also linear with the number of classes. 

The objective function of SLCE can be modified to accommodate a penalty term using $\ell_1$-norm to promote feature sparsity. In this setting, the algorithm could be used as a feature selector. The model can be easily extended to the semisupervised setting by adding the sample reconstruction term of PCA. With its current form, SLCE reconstructs class centroids from the samples in the same class. In reduced space, two classes may overlap if the class centroids are close in ambient space. Adopting a cost that also caters to separating the different classes in embedding space may benefit classification tasks. In the future, we will explore these ideas.

\acks{
We would like to acknowledge support for this
research from the National Science Foundation under award
NSF-ATD 1830676.}
\newpage
\appendix
\section*{Appendix A. Proof of $\Lambda$ being diagonal}
\label{diag_lambda}
The Lagrangian of rank $k$ projection,
\begin{equation}
\begin{aligned}
\mathcal{L}({A},\Lambda) = \|\tilde C-{A} {A}^T X \|_F^2 -tr(\Lambda ({A}^T{A}-I))
\end{aligned}
\label{equation:LCE_cost_k-dim_equ1_supply}
\end{equation}

For the rank $k$-projection, we have $k+k(k-1)/2$ number of constraints. The $k$-constraints are for each column of matrix A having unit norm, i.e. ${A_i}^TA_i = 1$. The $k(k-1)/2$-constraints are for each pair of ${A_i}^TA_j=0, i \ne j$, i.e., column $i$ and $j$ of $A$ are orthogonal to each other. For each constraints there will be a Lagrange multiplier as shown below,

\begin{equation}
\begin{aligned}
\mathcal{L}({A},\Lambda) = \|\tilde C-{A} {A}^T X \|_F^2 - \lambda_1 ({A_1}^TA_1 - 1) - \ldots - 
\lambda_k ({A_k}^TA_k - 1) - \\ \lambda_{k+1}({A_1}^TA_2 - 0) - \lambda_{k+1}({A_2}^TA_1 - 0) - \ldots - \lambda_{k(k-1)/2}(A_{k(k-2)/2}^TA_{k(k-1)/2} - 0) \\ - \lambda_{k(k-1)/2}(A_{k(k-1)/2}^TA_{k(k-2)/2} - 0)
\end{aligned}
\label{equation:LCE_cost_k-dim_equ2}
\end{equation}
We can incorporate all the constraints by using the trace of the matrix $\Lambda ({A}^T{A}-I)$. Note the constraints from $k+1$ to $k(k-1)/2$ are repeated twice because of the symmetry of inner product. i.e. ${A_i}^TA_j = {A_j}^TA_i$. Notice that the diagonal entries of $\Lambda$ will contain the first $k$ Lagrange multipliers ($\lambda_1 \ldots \lambda_k$) and the off-diagonal elements will contain the rest with $\Lambda_{i,j} = \Lambda_{j,i}$. It's clear that $\Lambda$ is a symmetric matrix and therefore we can do eigendecomposition of as follows, $\Lambda = V \Phi V^T$, where $V$ is a orthogonal matrix such that $V^TV = VV^T = I$, and $\Phi$ is a diagonal matrix which contains the eigenvalues. We can expand the terms in trace in Equation \ref{equation:LCE_cost_k-dim_equ1_supply} as follows,

\begin{equation}
\begin{aligned}
tr(\Lambda ({A}^T{A}-I)) = tr(V\Phi V^T({A}^T{A}-I))\\
tr(\Lambda ({A}^T{A}-I)) = tr(\Phi V^T({A}^T{A}-I)V)\\
tr(\Lambda ({A}^T{A}-I)) = tr(\Phi (V^T{A}^T{A}V-V^TIV))\\
tr(\Lambda ({A}^T{A}-I)) = tr(\Phi (V^T{A}^T{A}V-I))\\
tr(\Lambda ({A}^T{A}-I)) = tr(\Phi (M^TM-I))\;\;where\;\;M:=AV\\
\end{aligned}
\label{equation:LCE_cost_k-dim_equ3}
\end{equation}
Notice$A = MV^T$ and $AA^T = MM^T$. Now we can rewrite the cost function in Equation\ref{equation:LCE_cost_k-dim_equ1_supply} as follows,

\begin{equation}
\begin{aligned}
\mathcal{L}({M},\Phi) = \|\tilde C-{M} {M}^T X \|_F^2 -tr(\Phi ({M}^T{M}-I))
\end{aligned}
\label{equation:LCE_cost_k-dim_equ4}
\end{equation}

Comparing Equation \ref{equation:LCE_cost_k-dim_equ1_supply} with \ref{equation:LCE_cost_k-dim_equ4}, we can have a change of variable $M=AV$ and make the Lagrange multiplier as a diagonal matrix $\Phi$. At last we show that if $A$ is the optimal solution of Equation \ref{equation:LCE_cost_k-dim_equ1_supply}, then $M$ is also a solution. The solution of \ref{equation:LCE_cost_k-dim_equ1_supply} comes as a eigendecomposition problem in the form,
\begin{equation}
\begin{aligned}
( X \tilde C^T + \tilde C X^T - XX^T) {A} = {A} \Lambda
\end{aligned}
\label{equation:LCE_cost_k-dim_eq5}
\end{equation}

\begin{equation}
\begin{aligned}
( X \tilde C^T + \tilde C X^T - XX^T) {A} = {A} V\Phi V^T
\end{aligned}
\label{equation:LCE_cost_k-dim_eq6}
\end{equation}

\begin{equation}
\begin{aligned}
( X \tilde C^T + \tilde C X^T - XX^T) {A}V = {A} V\Phi
\end{aligned}
\label{equation:LCE_cost_k-dim_eq7}
\end{equation}

\begin{equation}
\begin{aligned}
( X \tilde C^T + \tilde C X^T - XX^T) M = M\Phi
\end{aligned}
\label{equation:LCE_cost_k-dim_eq8}
\end{equation}
It can be seen that the change of variable gives the same eigendecomposition problem. Hence we can write the Lagrange multiplier as a diagonal matrix $\Phi$.

\section*{Appendix B. Linearly dependency of the centroid matrix after mean subtraction}
\label{Liner_dependency}

$X \in \mathbb{R}^{d \times n}$ is the data matrix with $n$ samples where each sample $x_i \in \mathbb{R}^d$. $X$ has $M$ classes, where $I_j$, $|C_j|$, and $c_j$ are the index set, cardinality and the mean of $j^{th}$ class. We have the centroid matrix $\tilde C$ which contains the corresponding class centroid of $x_i$. Note, like $X$, $\tilde C \in \mathbb{R}^{d \times n}$ but $\tilde C$ has repeated entries. As $X$ is mean-subtracted, we can write,

\begin{equation}
\begin{aligned}
\sum_{i=1}^{n} x_i = 0
\end{aligned}
\label{equation:meansubtract_eq1}
\end{equation}

\begin{equation}
\begin{aligned}
\sum_{j \in I_1} x_j + \sum_{j \in I_2} x_j + \ldots + \sum_{j \in I_M} x_j = 0
\end{aligned}
\label{equation:meansubtract_eq2}
\end{equation}

\begin{equation}
\begin{aligned}
|C_1| \sum_{j \in I_1} \frac{1}{|C_1|} x_j + |C_2| \sum_{j \in I_2} \frac{1}{|C_2|} x_j + \ldots + |C_M|\sum_{j \in I_M} \frac{1}{|C_M|} x_j = 0
\end{aligned}
\label{equation:meansubtract_eq3}
\end{equation}

\begin{equation}
\begin{aligned}
|C_1| c_1 + |C_2| c_2 + \ldots + |C_M|c_m = 0
\end{aligned}
\label{equation:meansubtract_eq4}
\end{equation}

\begin{equation}
\begin{aligned}
- \frac{|C_1|}{|C_M|} c_1 - \frac{|C_2|}{|C_M|} c_2 - \ldots - \frac{|C_{M-1}|}{|C_M|}c_{m-1} = c_m
\end{aligned}
\label{equation:meansubtract_eq5}
\end{equation}

Equation \ref{equation:meansubtract_eq5} shows that $m^{th}$ class centroid is a linear sum of the rest. Therefore the matrix $\tilde C$ is linearly dependent and $rank(\tilde C) = M-1$

\bibliography{mybibfile}

\end{document}